# OntoSOC: Sociocultural Knowledge Ontology

Guidedi Kaladzavi[1], Papa Fary Diallo[2,3,] Kolyang[1], Moussa Lo[2]

[1]University of Maroua, Maroua, Cameroon
[2]University of Gaston Berger, UFR-SAT, Saint-Louis, Senegal
[3]University of Nice Sophia Antipolis, CNRS Nice, France

*Abstract.* This paper presents a sociocultural knowledge ontology (OntoSOC) modeling approach. OntoSOC modeling approach is based on Engeström's Human Activity Theory (HAT). That Theory allowed us to identify fundamental concepts and relationships between them. The top-down precess has been used to define differents sub-concepts. The modeled vocabulary permits us to organise data, to facilitate information retrieval by introducing a semantic layer in social web platform architecture, we project to implement. This platform can be considered as a « collective memory » and Participative and Distributed Information System (PDIS) which will allow Cameroonian communities to share an co-construct knowledge on permanent organized activities.

## 1 Introduction

Regarding the rhythm of the current cultural mixing, we can believe that in the long, culture of African people in particular may disappear due to its marginalization, its abandonment by the complicity of Africans themselves in favor of Western culture [1][2]. Those cultures are deteriorating and emptying of their meaning, their mellow content and values in many African peoples. Even the contents proposed by medias, educational systems and Internet are turned to the West side; this doesn't help African youth to know their culture. Yet, culture seems increasingly guide human activities. Some ways out seem identified : a permanent (re)education and of course internet. On the web, all topics are covered. It constitutes in that capacity, the most global source of information.

To refresh the memory of our citizens and to give a transparent view of opportunities (unknown infrastructures, etc.) and challenges (investments unequally distributed, etc.) through endogenous information (from involved actors), rather than external analysis, we project, the implementation of ontology-based web platform for sociocultural knowledge sharing and co-construction, in Cameroonian context. Here, sociocultural knowledge concept, concerns all forms of human knowledge: objects that compound the real world, facts and events [3].

Ontology is a organized conceptualization to produce a formal object, when relationships between concepts are semantic and formalized. It allows organizing data, improvement of information retrieval and automated indexation by introducing a semantic layer in the semantic web applications architecture. As such, ontologies are considered as knowledge representation tools transforming data into information and information into knowledge. The semantic web is an evolution of Web, through ontologies; it aims to allow software agents to understand that, the web pages content is not just a sequence of characters and pixels but intelligible information in the same way as humans.

The issue of this paper is to present our sociocultural knowledge ontology (OntoSOC) modeling approach. Which ontology will be used by our platform. OntoSOC modeling approach is based on Engeström's Human Activity Theory (HAT). HAT is a conceptual framework which the foundational concept is "activity", which is understood as purposeful, transformative, and developing interaction between actors ("subjects") and the world ("objects"). That Sociocultural ontology will allow us to identify which type of data could be shared on our platform.

This paper continues with background in which we present the related works to our ontology. Then, the third part presents our modeling approach which based on the Engeström's Human Activity Theory; we will explain how we reuse the related ontologies to solve the interoperability issue. In part four, we present how this ontology could be used through use cases and SPARQL queries. We end with a conclusion and perspectives for this work.

## 2 Related works

To our best knowledge only the Sociocultural Ontology [4][5] cover the sociocultural domain which has been developed in the Senegalese context. The modeling approach based on the first generation of HAT of Vygotsky. The model is organized around the "mediation" concept and based on the idea that, human actions are mediated by cultural, symbolic or physical artifacts that enable man to act on his environment. We agree that, the principle of authors on ontology should be more oriented on knowledge shared by communities rather than relationships between individuals. However, the translation of view done, considering the community as an atomic entity hides a set of information that we consider relevant in our context. It is the case of the internal dynamics (collaboration, interaction, actors and roles, etc.) of communities and the contextual nature (regulations, used tools, etc.) for organizing any activity. Not only, this information further enrich the knowledge base considered both as a "collective memory" and a standard PDIS on Cameroon but also allows conducting deep analyzes of communities and activities.

However, since the socio-cultural field is multidisciplinary, we reuse some related vocabularies which are:

- Foaf[1] : FOAF (Friend Of A Friend) project is a standard vocabulary that describes in RDF relationships between people while Wai[2] is its extension , which defines roles that people can play in community. We plan to model the cultural aspects of a community that essentially consists of people that FOAF and Wai allow us to model.
-
-Schema.org[3] is a collection of websites developers shared concepts in Microdatas, RDFa and JSON-LD formats to make the web pages content understandable by search engines like Google, Yahoo etc. DBpedia[4] is an academic and a community project for automatic data exploration from Wikipedia to propose a structured version in semantic web format of data. Both vocabularies cover a large part of socio-cultural aspects of locality and community that we intend to reuse in OntoSOC.

## 3 OntoSOC Modeling

Practically, ontological engineering does not propose a standardized method or general methodology for building ontologies [6]. Most of existing approaches adopt an iterative and incremental cycle. In our case, we used HAT based-methodology coupled with the top-down approach. We agree with Berger on the fact that "the reality is a social construction[5]". The universe is evolving. These changes are driven by groups of individuals through their various activities.To analyze and understand these changes and how they transform reality, many human activities models have been proposed. Among them, we have the Vygotsky's model [7] and Engeström's model [8]. Engeström defines four characteristics that allow us to model dynamism within communities and the contextual factor of each activity:

- Human activity must be represented by its simplest expression;
- Human activity must be parsed in terms of its dynamism and its historical transformation;
- Human activity should be considered as a contextual phenomenon, that is to say, as an ecosystem of relationships between people and their environment;
- Human activity should be interpreted as a phenomenon mediated by culture (more than a dialectical relationship).

---

[1] http://xmlns.com/foaf/spec/
[2] http://lov.okfn.org/dataset/lov/
[3] https://schema.org/
[4] http://dbpedia.org/

To achieve our concern which is modeling knowledge related to various activities organized daily in our localities we have chosen Engeström's model as formal framework basis of our modeling process.

### 3.1 Engeström Human Activity Model

The Engeström Human Activity Theory is known as third generation model. It points out clear distinction between individual and collective activity. It's produced according to the historical and cultural view of Activity. According to its author, it helps to have a holistic view of local change dynamics while considering culture and historicity [8]. The model has six poles (*Subject, Object, Tool, Rules, Community and Division of Labour*) as shown in Figure 1.

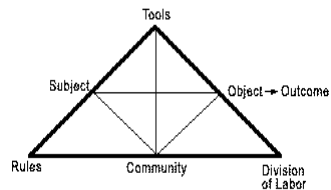

Fig. 1: Engeström Human Activity triangle

─ *Subject* : represents the individual chosen to analyze by the observer ;
─ *Object* : transformation of the environment by the activity (task to be performed, objective to be achieved);
─ *Tools* : materials or symbolic tools that mediate the activity ;
─ *Comm*unity : set of subjects (or subgroups) that share the same objects and differ thereby other communities;
─ *Division of Labour :* it considers both the horizontal distribution of actions among the subjects, community members and the vertical hierarchy or responsibilities and statutes;
─ *Rules* : they refer to standards, conventions, habits, etc. implicit and explicit that maintain and regulate the actions and interactions within the system.

### 3.2 Fundamental concepts and Relationships identification process

The Engeström Theory is characterized by individual and collective vision of the activity concept and Any HAT has meaning only in a social matrix (context). Thus, the "Collaborative persona" method applied to three use cases has helped us to analyze the different triads constituting the

---

[5] http://mipms.cnam.fr/servlet/com.univ.collaboratif.utils.LectureFichiergw?ID_FICHIER=1295877017861

model.This approach allowed us to identify all possible relationships between different poles. To illustrate, the *Rules-Object Community*-Triad extracted from the global triangle, allows us to write the following triples (Fig. 2):

- **is*RespectedBy*** *( Rules,Object) ;*
- **is*OrganisedBy*** *(Object, Community) ;*
- **is*RegulatedBy*** *(Community, Rules).*

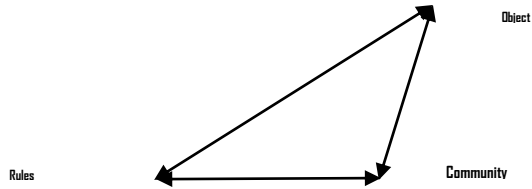

Fig. 2. Rules-Activity-Community Triad

The approach applied to twelve (12) triads within the model allows us to point out different uses of each pole from the three use cases that we designated respectively by case1[6], case2[7], case3[8] (see Table 1). In first case, the *community* pole was involved in seven (07) different triads, same for the other two, and at all, twenty one (21) implications of the concept in different triads.This process permitted us to decide the existence or otherwise, and semantics of the relationships between the different poles of the model. For example, in *Subject-Community-Tools* and *Tools-Community-Object* triads, a triplet *BelongsTo (Tools, Community)* could exist. Indeed, any tool used during the activity execution does not necessarily belongs to the community that organizes it, for that purpose, we decided not to consider the possible relationship between tools and community.

Otherwise, from *Rules-Community-Object* and *Subject-Object-Rules* triads, the overall semantics got out from used situations of *rules* concept have allowed to tie it to the Community *by isRegulated* relationship.

Table 1. implications of differents poles

|  | *Community* | *Object* | *Subject* | *Rules* | *Division of Labour* | *Tools* |
|---|---|---|---|---|---|---|
| **Case 1** | 7 | 7 | 7 | 3 | 3 | 3 |
| **Case 2** | 7 | 7 | 7 | 3 | 3 | 3 |
| **Case 3** | 7 | 7 | 7 | 3 | 3 | 3 |
| **Total** | 21 | 21 | 21 | 9 | 9 | 9 |

Finally ; after eliminating redundant triplets, we have the following triplets: **isUsedBy** (Tools, Subject), **isMemberOf** (Subject, Community), **isRegulatedBy** (Community, Rules), **isCreatedBy** (Division of Labour, Community), **plays**(Subject, Division of Labour), is**RealisedBy** (Division of Labour, Object), **isOrganisedBy**(Object, Community), **isLocatedIn**(Community, Locality), **isOccuredIn**(Object, Locality), **isBorderdBy** (Locality, Locality).

---

[6] Cultural community *Naakosenda* was engaged in cultural event

[7] *CDE-SAARE* was building the rural library

[8] *Club 2-0* organized holidays soccer tournament

The subjects and objects of these triplets above model the fundamental ontology concepts, predicates represent the different relationships between them. Table 2 shows the mapping carried out between the poles of the Activity model and the basic concepts of our ontology.

Table 2. Mapping between poles of Activity model and ontology concepts

| Poles of Engeström Model | fundamental ontology Concepts |
|---|---|
| Tools | Resource |
| Object | Activity |
| Subject | Individual |
| Rules | Regulations |
| Community | Community |
| Division of Labour | Role |

### 3.3 Composition of OntoSOC Concepts et Relationships.

The different concepts obtained by mapping represent fundamental classes of our ontology. We have, in all seven (07) upper-level concepts. Figure 3 illustrates the upper level concepts and relationships.

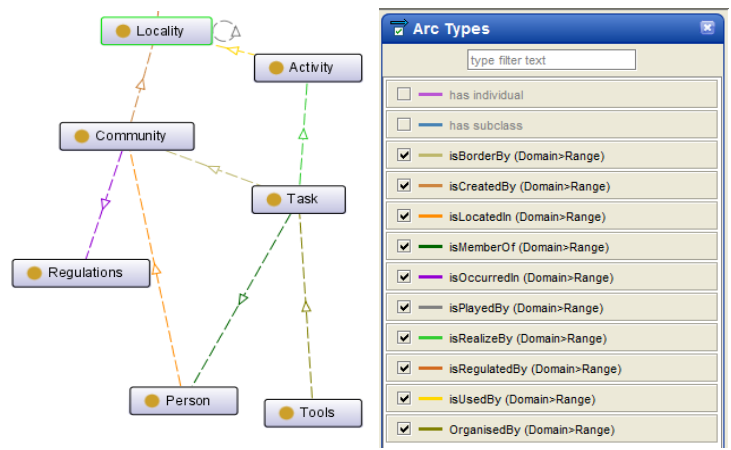

Fig. 3. OntoSOC upper level concepts and relationships

— *Community* : set of people who share a common values and interests;
— *Resource*: represents materials and symbolic tools used during activity running.
— *Régulations* : represent the different texts, laws that regulate communities;
— *Activity* : Any event organised by Community;
— *Individual/person* : Any Community member.
— *Locality* : Place (administrative) where activity is held ;
— *Role* : action, operation, played by individual or set of individuals during activity realization.

### Hierarchy of classes

Excepted, the *individual* concept, each concept has variant depth of sub-concepts going from 1 to 5. There are many approaches to develop hierarch of class : from top-down, bottom-up, to hybrid approach. The use of HAT which allowed us to have the basic concepts is the beginning of the top-down method. Thereafter, to better define the hierarchy, we intended to think up before making specifications. Which is the top-down development process that begins with a definition of the most general concepts in the domain and continues with sub-concepts specialization. Figure 4 shows the extract of a possible articulation between various levels generality of *Activity* concept.

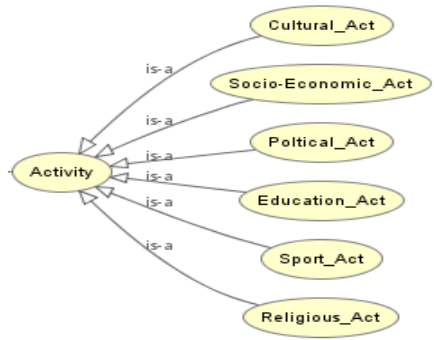

Fig.4. Activity concept Hierarchy

### 3.4 OntoSOC alignment

For semantic web, alignment is solution to the interoperability problem between heterogeneous ontologies developed and help us not to recreate that exists but only improve it. OntoSOC is an inter-domain vocabulary. It reuses some concepts of related ontologies. Figure 5 presents the mapping, done between related vocabularies and some OntoSOC concepts.

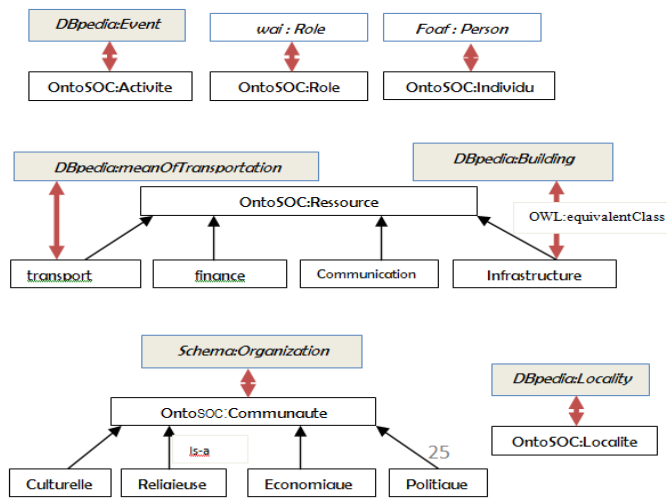

Fig.5. OntoSOC concepts alignment

## 4 SPARQL queries and populating
### 4.1 Populating

For editing, we used the ontology development tool "Protégé 4.3". Protégé is a platform of ontologies implementation and management, using tools for modeling different areas and knowledge-based applications with specific ontologies [9]. Its Populating was done with three use cases data according to "collaborative persona" approach. The cultural community called *Naakonda* engaged in organizing a cultural event in Mokolo locality, The CDE-SAARE[9] community was building a rural library in Kolara and sport community called Club 2-0 organizing a soccer tournament holidays in « College L'Espoir » of Maroua.

---

[9] www.cde-saare.de

The persona method is a modeling strategy used by software architects. This idea was introduced by Alan Cooper, software designer[10]. In Software Engineering, this approach is called « Goal-directed design ». Personas are fictional personifications which represent realistic individual persons. In our case, we used "collaborative persona". The Advanced form of individual persona. It is suitable for collaborative, participation and interaction context in communities or groups [11]. By creating some humanization of collaborations within three use cases activities, we used (for populating) data-instances of OntoSOC concepts within twelve (12) triads that make up the HAT. The following figure shows that *Naakosenda* member called *Tangoche* participating in the cultural event. It illustrates instances used during populating.

### 4.2 SPARQL queries

A SPARQL endpoint is available in the Protégé environment. It provides to build queries on OWL graph of OntoSOC.The use of some predicates as owl: disjointWith, owl:equivalentClass, owl:hasValue, respectively helped us to return (if any) the concepts are disjoint, concepts that are equivalent and finally the exact values of properties (who own one). We focused on various activities organized in a specific locality, and the tools used for the cultural event. The following query returns to any given community, organized activities, resources used and the roles played by actors.

```
PREFIX rdf: <http://www.w3.org/1999/02/22-rdf-syntax-ns#>
PREFIX owl: <http://www.w3.org/2002/07/owl#>
PREFIX xsd: <http://www.w3.org/2001/XMLSchema#>
PREFIX rdfs: <http://www.w3.org/2000/01/rdf-schema#>
PREFIX OntoSOC: <http://maroua-univ/ns/ontosoc#>
SELECT ?Communities ?Activity ?task ?person ?tools
WHERE {?task OntoSOC:isUsedBy ?tools
            OPTIONAL { ?Activity OntoSOC:isRealizeBy ?task }
            OPTIONAL { ?task OntoSOC:isPlayedBy ?person }
            OPTIONAL { ?task OntoSOC:isCreatedBy ?Communities }
} ORDER BY  ?Communities
```

### 5   Conclusion and perspectives

In this paper, we presented a modeling approach of sociocultural knowledge ontology named OntoSOC. To get there, we simulated the Engeström Human Activity Theory to identify upper-level ontology concepts and relationships between them. The developed ontology is considered as an improved version of ontology developed by [5], But in the Cameroonian context. To fix the interoperability problem we established mapping between OntoSOC concepts and those related.We populated our ontology with three use cases and applied some SPARQL tests. Certainly, the three cases are far to be representative, but, their data helped us to eliminate or explain some inconsistencies.

It should be noted that the elimination of redundant triplets was done empirically. We have no guarantee of reaching the minimal coverage. Nevertheless, the reduction rate is considerable, about 60%. In addition, the notion of sociocultural knowledge is not easy to be well-mastered: it can be specific to a problem, generally to a domain, deep; surface, accurate, uncertain, imprecise or incomplete. However, OntoSOC informs us about data type to be shared and co-constructed into our platform.

In following the work, to design the best architecture of our platform we focus on domain ontology modeling. That will allow us to point out if all poles of TAH may still exist depending on community or activity type.